\documentclass{article}

\usepackage[final]{neurips_2024_ml4ps}

\usepackage[final]{neurips_2024_ml4ps}
\usepackage{booktabs}       
\usepackage{amsfonts}       
\usepackage{nicefrac}       
\usepackage{microtype}      
\usepackage{xcolor}         
\usepackage{tabularray}
\usepackage{rotating}
\setcitestyle{round}

\title{Uncertainty Quantification
for Surface Ozone Emulators using Deep Learning}

\author{%
  Kelsey Doerksen \\
  OATML \\
  University of Oxford\\
  \texttt{kelsey.doerksen@cs.ox.ac.uk} \\
  \And
  Yuliya Marchetti \\
  Jet Propulsion Laboratory \\
  California Institute of Technology
  \And
  Steven Lu \\
  Jet Propulsion Laboratory \\
  California Institute of Technology
  \And
  Kevin Bowman \\
  Jet Propulsion Laboratory \\
  California Institute of Technology
  \And
  James Montgomery \\
  Jet Propulsion Laboratory \\
  California Institute of Technology\\
  \And
  Kazuyuki Miyazaki \\
  Jet Propulsion Laboratory \\
  California Institute of Technology
  \And
  \And
  Yarin Gal \\
  OATML \\
  University of Oxford
  \And
  Freddie Kalaitzis \\
  OATML \\
  University of Oxford\\
}

\begin{document}

\maketitle

\begin{abstract}
Air pollution is a global hazard, and as of 2023, 94\% of the world's population is exposed to unsafe pollution levels. Surface Ozone (O3), an important pollutant, and the drivers of its trends are difficult to model, and traditional physics-based models fall short in their practical use for scales relevant to human-health impacts. Deep Learning-based emulators have shown promise in capturing complex climate patterns, but overall lack the interpretability necessary to support critical decision making for policy changes and public health measures. We implement an uncertainty-aware U-Net architecture to predict the Multi-mOdel Multi-cOnstituent Chemical data assimilation (MOMO-Chem) model's surface ozone residuals (bias) using Bayesian and quantile regression methods. We demonstrate the capability of our techniques in regional estimation of bias in North America and Europe for June 2019. We highlight the uncertainty quantification (UQ) scores between our two UQ methodologies and discern which ground stations are optimal and sub-optimal candidates for MOMO-Chem bias correction, and evaluate the impact of land-use information in surface ozone residual modeling.
\end{abstract}

\section{Introduction}
The latest estimate of the global burden of disease indicates that ambient air pollution, including fine particulate matter and ozone, contributes to an estimated 5.2 million deaths each year [1]. To effectively assess the level of human exposure to air pollution, accurate and timely estimates are necessary. MOMO-Chem, a state-of-the-art data assimilation framework for modeling surface ozone, has shown significant progress in capturing large-scale ozone estimates [2]. The MOMO-Chem uses four Chemical Transport Models, namely GEOS-Chem [3], AGCM-CHASER [4], MIROC-Chem, and MIROC-Chem-H [5] that utilize multi-constituent satellite data and an ensemble Kalman filter to propagate observational information in time and space from a limited number of observed species to a wide range of chemical components. However, the model struggles with finer-scale ozone analysis due to large systematic estimation errors (hereby referred to as bias), leading to a limited understanding of surface-level air quality and consequentially, its health impacts. In MOMO-Chem, bias can be driven by a mixture of poorly constrained processes including atmospheric chemistry, anthropogenic emissions and planetary boundary layer dynamics. AI-powered climate model emulators have grown in popularity for their relative speed and computational power to capture complex climatological processes compared to physics-based models. A neural network approach for a statistical Regional Climate Model-emulator to learn the relationship between large-scale predictors and a local surface variables has been developed [6], and ClimateBench provides a benchmarking framework for AI climate emulation, of which a range of emissions and concentrations of carbon dioxide, methane and aerosols is provided [7]. To transform estimates into decisions, and for acceptance by the scientific community, these models must provide a measurement of confidence. Uncertainty Quantification provides metrics by which we can evaluate AI-generated estimates that gives confidence to users, but there is currently no "one-size fits all" UQ methodology widely accepted by the scientific community. Some work has been done to utilize UQ in a scientific setting in selective classification to support clinical trials through withholding predictions in uncertain scenarios [8], and in a climate context, selective prediction based on predictive uncertainty for wind speed estimation of tropical cyclones with Deep Neural Networks [9]. Our downstream task is similar to selective prediction, in that we would like to identify ground stations candidates for bias correction by domain experts using the predictive uncertainty as a metric for easy/difficult candidates to correct. We explore and compare two UQ methodologies using a Bayesian approximation and quantile-based method via Monte-Carlo (MC) Dropout and Conformalized Quantile Regression (CQR) to discern the differences in UQ for modeling surface ozone bias in our Deep Learning air quality emulator.  We investigate which ground-based air quality stations are the easiest and most difficult to model, and explore model performance with and without the integration of additional land-use information via Google Earth Engine satellite products to targeting surface ozone bias. We further extend our predictions to extrapolate predictions beyond the ground station coverage in North America. Our work aims to answer the following questions on modeling surface ozone bias in our Deep Learning emulator:
\begin{enumerate}
    \item Do we observe spatial consistencies in UQ techniques?
    \item Does the addition of land-use information improve model performance?
    \item Which ground stations do we struggle/excel at capturing bias?
\end{enumerate}

\section{Dataset}
\textbf{Input.} Our input image size ranges for the two geographic extents; 31x49 for North America and 31x27 for Europe. We investigate two feature space setups, a 28-channel input of only MOMO-Chem features selected by atmospheric science domain expert insight, consisting of: BrOx, C\textsubscript{10}H\textsubscript{16}, C\textsubscript{2}H\textsubscript{6}, C\textsubscript{3}H\textsubscript{6}, C\textsubscript{5}H\textsubscript{8}, CH\textsubscript{2}O, H\textsubscript{2}O\textsubscript{2}, HNO\textsubscript{3}, HO\textsubscript{2}, N\textsubscript{2}O\textsubscript{5}, NH\textsubscript{3}, OH, PAN, CO, COflux, NO, NO\textsubscript{2}, NOxFlux, precipitation, surface pressure, water vapour, long and short wave radiation, outgoing longwave radiation, temperature, cloud cover, and wind direction, and a 51-channel input of combined MOMO-Chem and Google Earth Engine (GEE) extracted satellite data products. GEE data was extracted and processed using the airPy package\footnote{https://github.com/kelsdoerksen/airPy} and includes the mode, variance, and percent coverage per land class per grid from the MODIS Land Cover Yearly product and the variance, maximum, minimum and average from the Global Population Density product [10-11]. All data is rasterized to a 11.1kmx11.1km resolution.

\begin{figure}[htbp]
  \centering{
\includegraphics[width=0.5\textwidth]{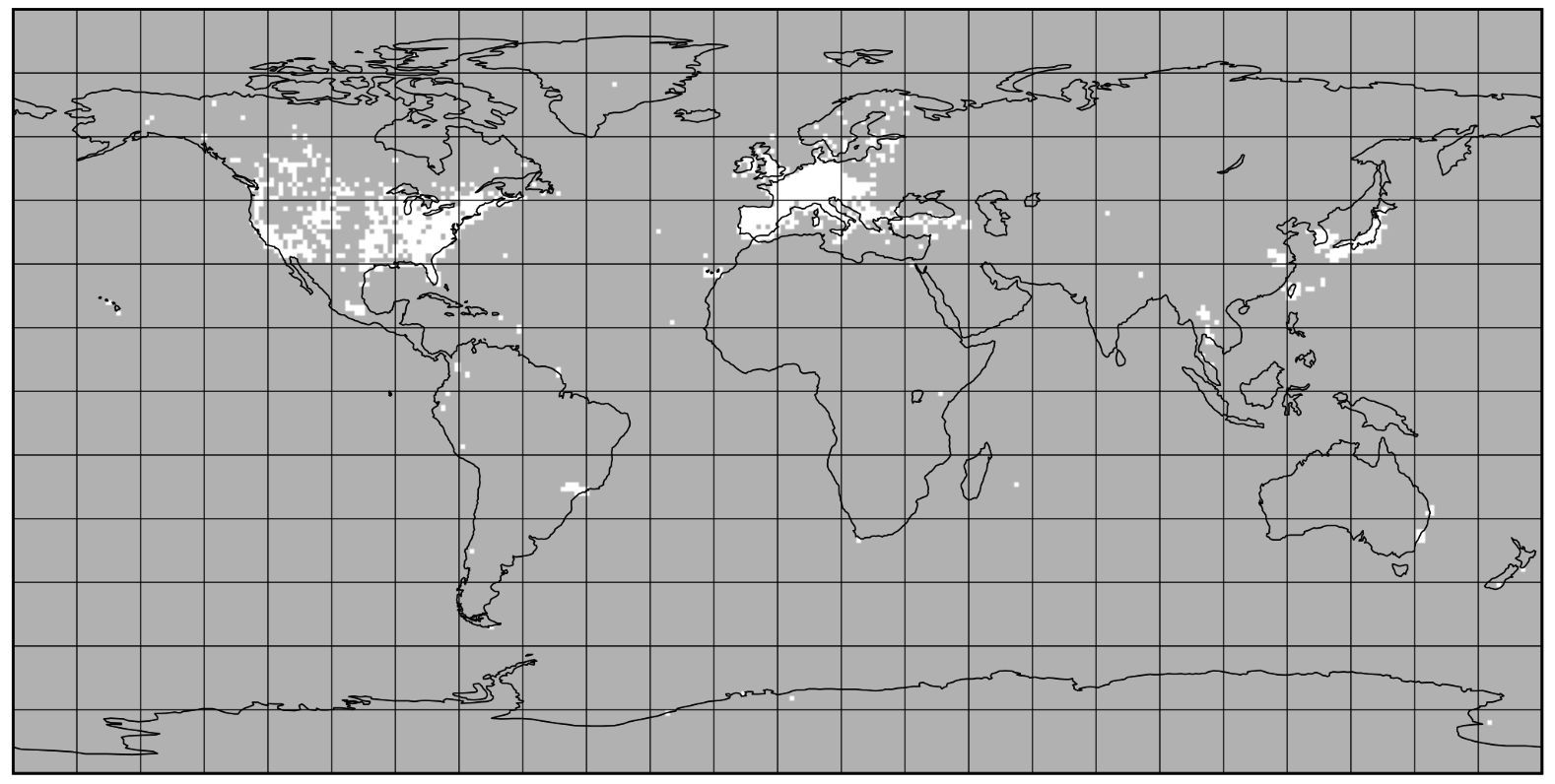}}
  \caption{TOAR Ground station coverage. White represents TOAR coverage, gray no coverage.}
  \label{fig:toar}
\end{figure}

\textbf{Target.} Our target is surface ozone bias, represented by the residuals of the difference between the MOMO-Chem 8-hour ozone data and our ground truth data of 8-hour daytime surface ozone measurements from the Tropospheric Ozone Assessment Report (TOAR) database, which provides up-to-date, global, near surface ozone measurements [12]. Though the TOAR network provides one of the best global databases or ground-based surface ozone measurements, spatial coverage is highly concentrated over North America and Europe as shown in Figure \ref{fig:toar}, prompting our selection of a regional-model approach.

\section{Methodology}
\textbf{Experimental Setup.}  We have a dataset \begin{math}D\end{math} which contains $(x_i, y_i)$ where $x_i$ in $X$ is multi-channel array of chemical and physical information describing the Earth's atmosphere and surface conditions about a location $I$ and $y_i$ in $Y$ is the MOMO-Chem model bias. We train individual models for North America and Europe extents, respectively. Models are trained on daily samples from 2005-2018 June using a 90/10 train/validation split. We test performance on a held-out test set of daily samples from June 2019. We leverage a U-Net architecture with Adam optimization, constant learning rate of 1e-3 and train for 200 epochs with a dropout rate of 0.1. Our MC-Dropout model uses Negative Log likelihood loss, and our CQR model uses Quantile loss. For our CQR model, we set our target coverage percent to 90. Our CQR model therefore has three predictive heads targeting 0.05, 0.5, and 0.95 quantiles to generate the 90th percentile predictive interval and the point prediction (using quantile 0.5). We run experiments with 5 random seeds and report average scores and variance.

\textbf{Uncertainty Quantification} We employ two deep learning UQ methodologies, a Bayesian approximation approach via Monte-Carlo  Dropout, and a quantile-based method via Conformalized Quantile Regression. \textbf{Monte-Carlo Dropout}, or MC-Dropout, utilizes dropout as a
Bayesian approximation during inference to compute the prediction uncertainty [13].
\textbf{Conformalized Quantile Regression} generates lower and upper bound prediction intervals between which the target variable lies with high probability. The uncertainty in the prediction of the target variable is quantified by the length of the prediction interval, calculated by the difference of the upper and lower prediction bounds [14]. This process is achieved by splitting our training set into halves representing a training and calibration set.

\textbf{Evaluation.} We evaluate the incorporation of land-use data on model performance by comparing RMSE scores and UQ metrics (interval length, epistemic uncertainty) with 28 vs 51-channel inputs. We highlight epistemic uncertainty from MC-Dropout as this can be reduced with including more information [15]. We further evaluate our UQ methods by spatially plotting the averaged scores to observe, if any, visible signatures of high and low UQ regions.

\section{Results}
\begin{table}[htbp]
\centering
\caption{Comparison of averaged model results using Conformalized Quantile Regression (CQR) and Monte-Carlo Dropout (MCD) uncertainty quantification for North America (NA), Europe (EU) models for 28 and 51 Features (Feats). All results are averaged for the 2019 June test set. Interval refers to CQR predictive interval length. RMSE reported with variance across 5 random seeds.}
\label{tab:results}
\resizebox{\linewidth}{!}{
\begin{tblr}{
  row{3} = {c},
  row{4} = {c},
  row{5} = {c},
  row{7} = {c},
  row{8} = {c},
  row{9} = {c},
  cell{1}{2} = {c},
  cell{1}{3} = {c},
  cell{1}{4} = {c},
  cell{1}{5} = {c},
  cell{1}{6} = {c},
  cell{1}{7} = {c},
  cell{1}{8} = {c},
  cell{1}{9} = {c},
  cell{1}{10} = {c},
  cell{2}{1} = {r=4}{},
  cell{2}{2} = {r=2}{c},
  cell{2}{3} = {c},
  cell{2}{4} = {c},
  cell{2}{5} = {c},
  cell{2}{6} = {c},
  cell{2}{7} = {c},
  cell{2}{8} = {c},
  cell{2}{9} = {c},
  cell{2}{10} = {c},
  cell{4}{2} = {r=2}{},
  cell{6}{1} = {r=4}{},
  cell{6}{2} = {r=2}{c},
  cell{6}{3} = {c},
  cell{6}{4} = {c},
  cell{6}{5} = {c},
  cell{6}{6} = {c},
  cell{6}{7} = {c},
  cell{6}{8} = {c},
  cell{6}{9} = {c},
  cell{6}{10} = {c},
  cell{8}{2} = {r=2}{},
  hline{2} = {-}{},
  hline{6} = {2-10}{},
}
             & \textbf{UQ} & \textbf{Feats} & \textbf{RMSE}       & \textbf{Max Interval} & \textbf{Min Interval} & \textbf{Avg Interval} & \textbf{Max Epistemic} & \textbf{Min Epistemic} & \textbf{Avg Epistemic} \\
\begin{sideways}\textbf{NA }\end{sideways} & CQR         & 28             & 18.40 $\mp$ 1.14          & 53                    & 28                    & 40                    & -                      & -                      & -                      \\
             &             & 51             & 17.41 $\mp$ 
 0.16          & 50                    & 30                    & 40                    & -                      & -                      & -                      \\
             & MCD         & 28             & \textbf{10.76 $\mp$  0.05} & -                     & -                     & -                     & 192                    & 14                     & 88                     \\
             &             & 51             & 10.88  $\mp$
 0.29         & -                     & -                     & -                     & 187                    & 15                     & 83                     \\
\begin{sideways}\textbf{EU }\end{sideways} & CQR         & 28             & 16.50 $\mp$  0.06          & 54                    & 27                    & 37                    & -                      & -                      & -                      \\
             &             & 51             & 15.93 $\mp$
 1.23          & 56                    & 29                    & 36                    & -                      & -                      & -                      \\
             & MCD         & 28             & 10.52  $\mp$ 
 0.17         & -                     & -                     & -                     & 67                     & 4                      & 22                     \\
             &             & 51             & \textbf{10.10 $\mp$  0.13} & -                     & -                     & -                     & 79                     & 4                      & 27                     
\end{tblr}
}
\end{table}

\begin{figure}[htbp!]
  \centering{
\includegraphics[width=1\textwidth]{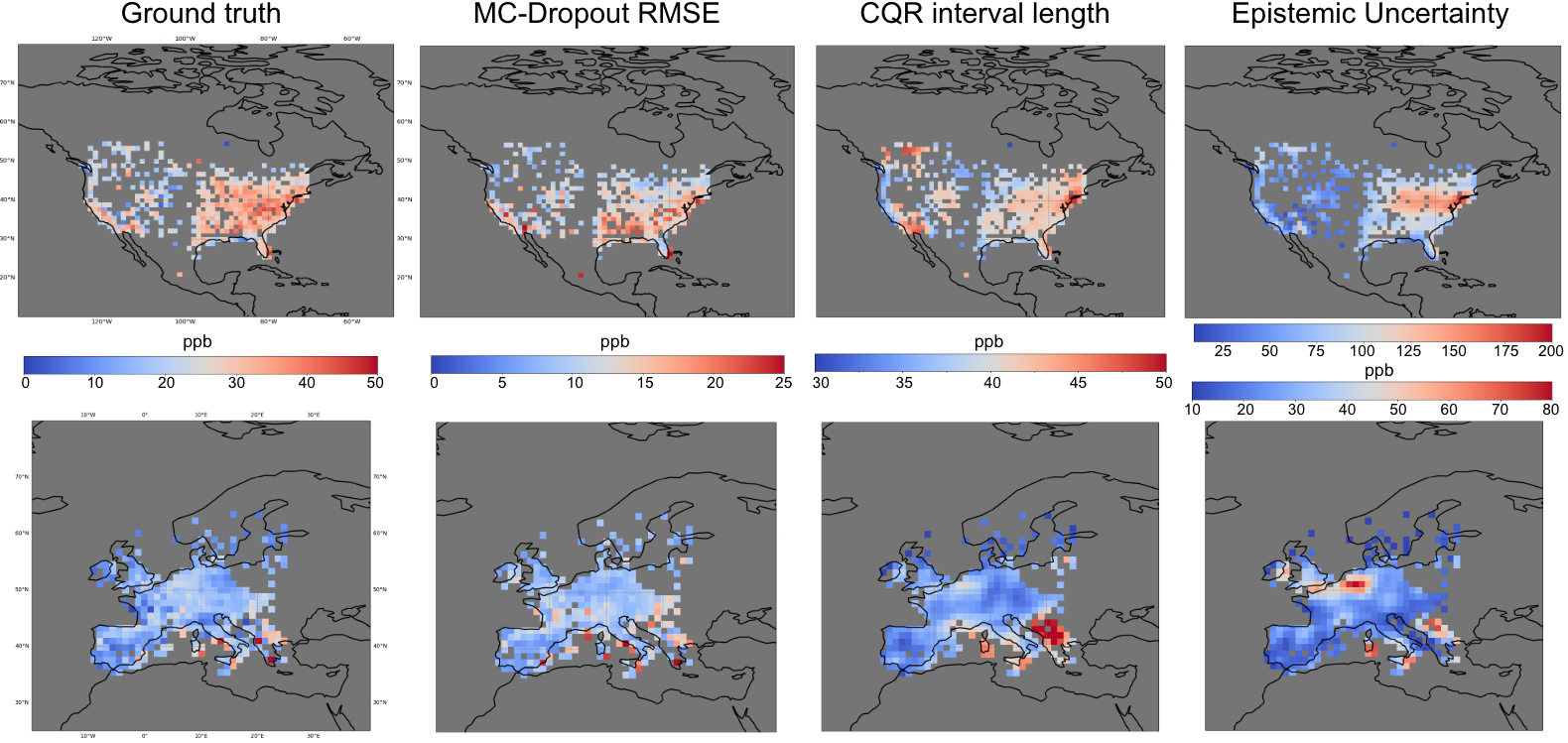}}
  \caption{51-channel MC-Dropout and CQR averaged model results for June 2019 Test set. Top: North America. Bottom: Europe. From left to right: Ground truth (ppb), MC-Dropout RMSE score (ppb), CQR average prediction interval length, MC-Dropout Epistemic Uncertainty. Colourbar scaling varies per metric, with shared scale represented by a single colourbar, and two colourbars for epistemic uncertainty corresponding to North America (top) and Europe (bottom) respectively. Colourbars move from dark blue (lower values) to dark red (higher values).}
  \label{fig:spatialresults}
\end{figure}

Table \ref{tab:results} summarizes model performance for June 2019. Referring to Question 2: Does the inclusion of land-use information improve model performance, we see small improvements in RMSE score for both the MC-Dropout and CQR models in Europe and CQR for North America, with no RMSE improvement in the MC-Dropout model in North America (28-channel RMSE slightly lower than 51-channel). For CQR, the averaged interval length does not show any discernable difference between the number of channels in the input space for North America or Europe. In North America, our MC-Dropout method shows a decrease in the averaged and maximum epistemic uncertainty with greater input features. In Europe, we see that increasing the number of input features increases the epistemic uncertainty, which could be caused due to feature correlation resulting in model confusion. Figure \ref{fig:spatialresults} depicts spatial maps of the averaged scores for the MC-Dropout and CQR methods in the North America and Europe models for June 2019. In North America, we see consistency in higher uncertainty scores between both methodologies on the Eastern coast of the continent. This pattern is also seen in the larger RMSE scores in the same region that correspond to high ground truth bias. Areas of high bias are hard to capture by the model in Europe, as referenced by the similarities in high bias and high RMSE regions (South-East). Again, we see consistency in our UQ methods for spatially identifying regions of high uncertainty in Europe, this time in the South-Eastern portion of the continent. We find differences between the MC-Dropout and CQR-identified high relative UQ regions across the North Western portion of Europe.

\begin{figure}[htbp!]
  \centering{
\includegraphics[width=0.7\textwidth]{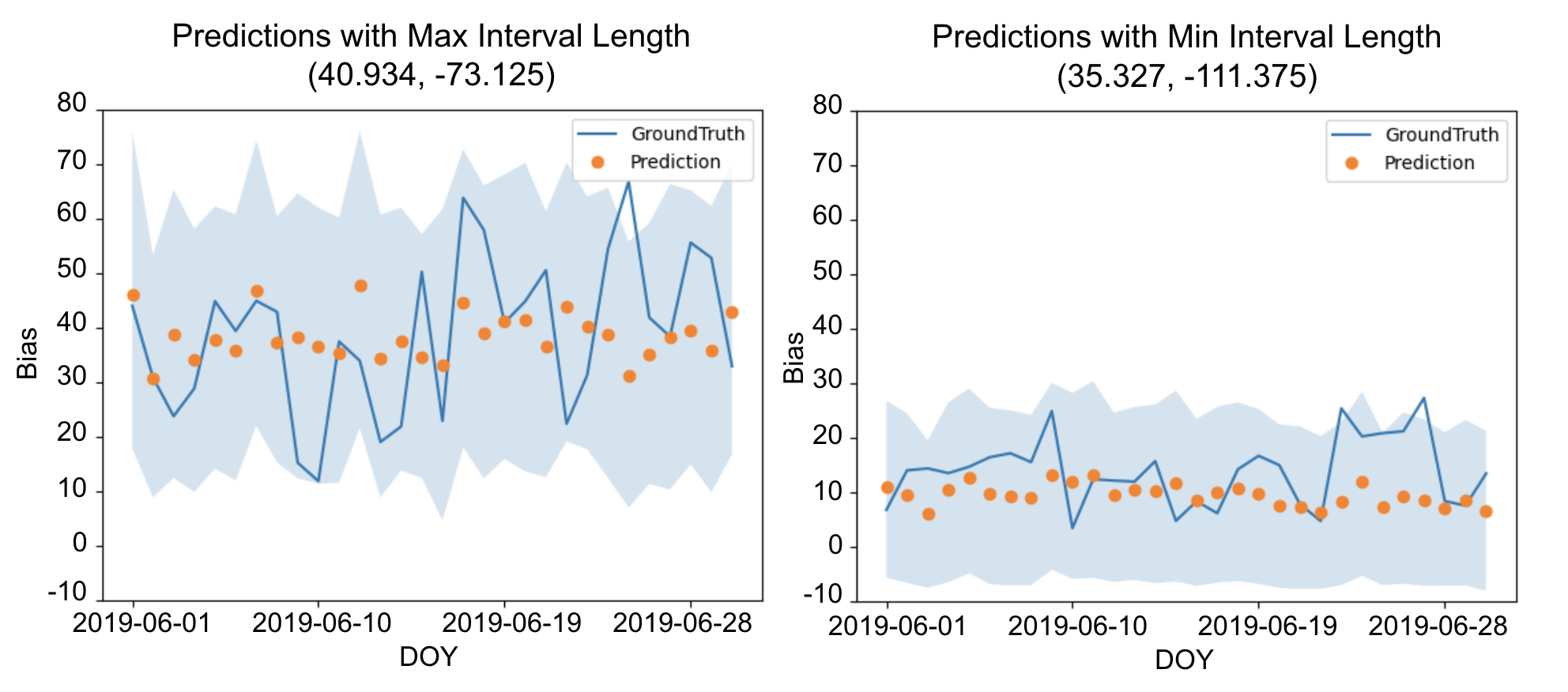}}
\caption{Time series of ground truth and CQR predictions for the ground station with, on average, the largest and smallest prediction interval length, North America. Orange points are model prediction, blue time series ground truth, light blue bands CQR prediction interval length. Bias in ppb.}
  \label{fig:minmax}
\end{figure}

We further investigate the ground stations which yielded the largest and smallest UQ scores and plot their time series, shown in Figure \ref{fig:minmax}. We selected the maximum UQ ground station candidate at latitude, longitude (40.934, -73.125) which yielded the highest UQ score in 2/5 seeds for CQR. We selected the minimum ground station candidate at latitude, longitude (35.327, -111.375) which yielded the smallest UQ score in 1/5 seeds for CQR and 1/5 seeds for MCD. At our max location (East coast), the ground truth signal is very noisy throughout the month of June, which our model struggles to capture (centering on the mean), resulting in large prediction intervals and an average (CQR model) RMSE of 14.36. At our minimum location (South-West), there is much less temporal variation in the ground truth signal, which creates an easier target for our models to estimate with an average (CQR model) RMSE of 7.18.

\begin{figure}[htbp!]
  \centering{
\includegraphics[width=1\textwidth]{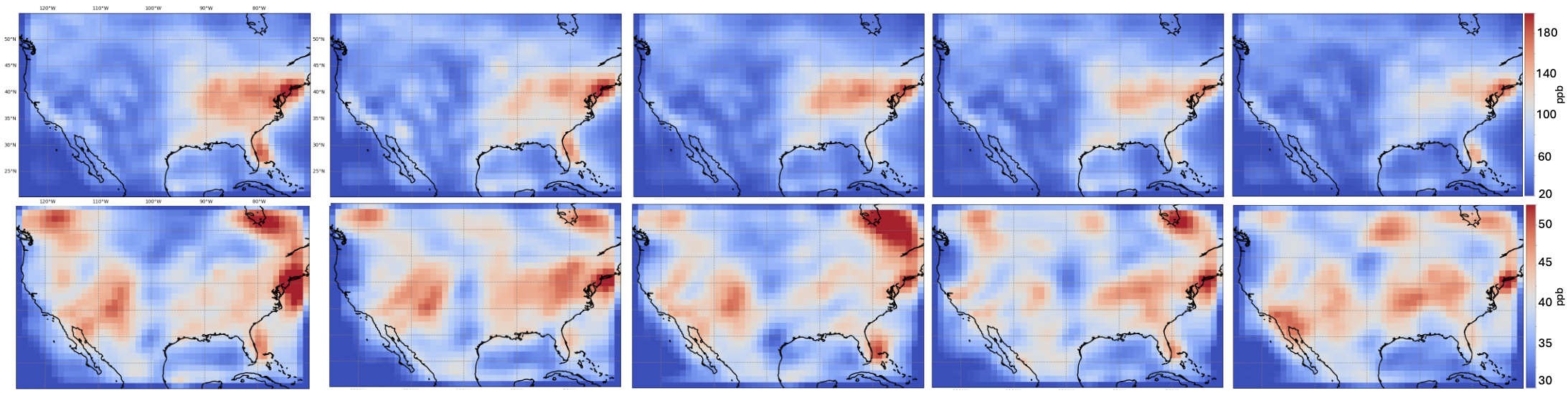}}
\caption{MC-Dropout epistemic uncertainty (top) and Conformalized Quantile Regression interval length (bottom) spatial extrapolation from model predictions using 28-channel feature space (left to right) June 1, 7, 15, 21, and 30th, 2019.}
  \label{fig:spatialextrap}
\end{figure}

\textbf{Spatial Extrapolation.} Figure \ref{fig:spatialextrap} shows a spatio-temporal comparison of the predictive uncertainty for the 28-channel feature space for the MC-Dropout (top) and CQR (bottom) techniques for June 1st, 7th, 15th, 21st, and 30th, 2019, respectively. We see that there is consistency among the techniques in identifying the East Coast of the United States as a high UQ region. The CQR model (bottom) shows high UQ on both coasts of the continent which is not reflected in the MC-Dropout method. There is a distinctive difference in the distribution of UQ throughout the month of June shown by both methods; in MCD we see the UQ at the beginning of the month over a larger portion of the East Coast than the end of the month, and with CQR a variability in the spatial scale and intensity of UQ in the North-East of the continent.

\section{Conclusion}
We presented a first analysis of uncertainty quantification for surface ozone bias estimation using quantile regression and Bayesian approximaton methods. Both techniques showed overlap in the spatial patterns of high and low UQ regions of North America and Europe that match with regions of high bias, but differed in their extrapolated results outside of ground station coverage, and exhibit a temporal variation of UQ over the month of June, prompting future studies in this area. We find that the inclusion of GEE data to our emulator improves epistemic uncertainty in North America, but does not otherwise show significant improvement. In future work, we aim to further investigate temporal UQ patterns across techniques, extending our work to a global scale, and include additional UQ methods to create a benchmark of UQ for surfaze ozone bias modeling.

\section{Acknowledgements}
KD acknowledges funding from EPSRC Centre for Doctoral Training in Autonomous Intelligent
Machines and Systems (Grant No. EP/S024050/1) and co-funding from the Oxford- Singapore
Human-Machine Collaboration Programme, supported by a gift from Amazon Web Services. A portion of this research was carried out at the Jet Propulsion Laboratory, California Institute of Technology, under a contract with the National Aeronautics and Space Administration (80NM0018D0004).
© 2024. All rights reserved.


\section*{References}
[1] Wei Huang, Hongbing Xu, Jing Wu, Minghui Ren, Yang Ke, and Jie Qiao. Toward cleaner air and better health: Current state, challenges, and priorities. Science, 385(6707):386–390, 2024. doi: 10.1126/science.adp7832.

[2] K. Miyazaki, K. W. Bowman, K. Yumimoto, T. Walker, and K. Sudo. Evaluation of a multi-model, multi-constituent assimilation framework for tropospheric chemical reanalysis. Atmospheric Chemistry and Physics, 20(2):931–967, 2020. doi: 10.5194/acp-20-931-2020.

[3] Isabelle Bey, Daniel J. Jacob, Robert M. Yantosca, Jennifer A. Logan, Brendan D. Field, Arlene M. Fiore, Qinbin Li, Honguy Y. Liu, Loretta J. Mickley, and Mar-
tin G. Schultz. Global modeling of tropospheric chemistry with assimilated meteo-
rology: Model description and evaluation. Journal of Geophysical Research: Atmo-
spheres, 106(D19):23073–23095, 2001. doi: https://doi.org/10.1029/2001JD000807.

[4] Kengo Sudo, Masaaki Takahashi, Jun-ichi Kurokawa, and Hajime Akimoto. Chaser: A global chemical model of the troposphere 1. model description. Journal of Geophysical Research: Atmospheres, 107(D17):ACH 7–1–ACH 7–20, 2002. doi: https://doi.org/10.1029/2001JD001113.

[5] S. Watanabe, T. Hajima, K. Sudo, T. Nagashima, T. Takemura, H. Okajima, T. Nozawa, H. Kawase, M. Abe, T. Yokohata, T. Ise, H. Sato, E. Kato, K. Takata, S. Emori, and M. Kawamiya. Miroc-esm 2010: model description and basic results of cmip5-20c3m experiments. Geoscientific Model Development, 4(4):845–872, 2011. doi: 10.5194/gmd-4-845-2011.

[6] A. Doury, S. Somot, and S. et al Gadat. Regional climate model emulator based on deep learning: concept and first evaluation of a novel hybrid downscaling approach. Climate Dynamics, 60: 1751–1779, 2023. doi: 10.1007/s00382-022-06343-9.

[7] D. Watson-Parris, Y. Rao, D. Olivi ´e, Ø. Seland, P. Nowack, and et al. Camps-Valls, G. Climatebench v1.0: A benchmark for data-driven climate projections. Journal of Advances in Modeling Earth Systems, (14), 2022. doi: 10.1029/2021MS002954.

[8] Lu Y, Chen T, Hao N, Van Rechem C, Chen J, and Fu T. Uncertainty quantification and interpretability for clinical trial approval prediction. Health Data Sci, 2024. doi: 10.34133/hds.0126.

[9] Nils Lehmann, Nina Maria Gottschling, Stefan Depeweg, and Eric Nalisnick. Uncertainty aware tropical cyclone wind speed estimation from satellite data, 2024. URL: https://arxiv.org/abs/2404.08325.

[10]  D. Sulla-Menashe and M.A. Friedl. Mcd12q1 modis/terra+aqua land cover type yearly l3 global 500m sin grid v006.

[11] CIESIN. Gridded population of the world, version 4 (gpwv4): Population density, 2018.

[12]  M.G. Schultz, Sabine Schr ¨oder, Olga Lyapina, Owen R. Cooper, and Ian Galbally. Tropospheric ozone assessment report: Database and metrics data of global surface ozone observations. Elementa: Science of the Anthropocene, (5:58), 2017. doi: 10.1525/elementa.244.

[13] Yarin Gal and Zoubin Ghahramani. Dropout as a bayesian approximation: Representing model uncertainty in deep learning. In Maria Florina Balcan and Kilian Q. Weinberger, editors, Proceedings of The 33rd International Conference on Machine Learning, volume 48 of Proceedings of Machine Learning Research, pages 1050–1059, New York, New York, USA, 20–22 Jun 2016. PMLR

[14] Yaniv Romano, Evan Patterson, and Emmanuel J. Cand `es. Conformalized quantile regression, 2019. URL https://arxiv.org/abs/1905.03222.

[15] Eyke H ¨ullermeier and Willem Waegeman. Aleatoric and epistemic uncertainty in machine learning: A tutorial introduction. CoRR, abs/1910.09457, 2019. URL
http://arxiv.org/abs/1910.09457.

\end{document}